\documentclass{article}

\usepackage[preprint]{neurips_2025}

\usepackage[utf8]{inputenc} 
\usepackage[T1]{fontenc}    
\usepackage{hyperref}       
\usepackage{url}            
\usepackage{booktabs}       
\usepackage{amsfonts}       
\usepackage{nicefrac}       
\usepackage{microtype}      
\usepackage{xcolor}         
\usepackage{multirow}
\usepackage{graphicx}  
\usepackage{amsmath}
\usepackage{array}

\title{PositionOCR: Augmenting Positional Awareness in Multi-Modal Models via Hybrid Specialist Integration}

\author{%
  Chen Duan \And
  Zhentao Guo \And
  Pei Fu \And
  Zining Wang \And
  Kai Zhou \And
  Pengfei Yan
}

\begin{document}

\maketitle
\begin{abstract}
In recent years, Multi-modal Large Language Models (MLLMs) have achieved strong performance in OCR-centric Visual Question Answering (VQA) tasks, illustrating their capability to process heterogeneous data and exhibit adaptability across varied contexts. However, these MLLMs rely on a Large Language Model (LLM) as the decoder, which is primarily designed for linguistic processing, and thus inherently lacks the positional reasoning required for precise visual tasks, such as text spotting and text grounding. Additionally, the extensive parameters of MLLMs necessitate substantial computational resources and large-scale data for effective training. Conversely, text spotting specialists achieve state-of-the-art coordinate predictions but lack semantic reasoning capabilities. This dichotomy motivates our key research question: Can we synergize the efficiency of specialists with the contextual power of LLMs to create a positionally-accurate MLLM? To overcome these challenges, we introduce PositionOCR, a parameter-efficient hybrid architecture that seamlessly integrates a text spotting model’s positional strengths with an LLM’s contextual reasoning. Comprising 131M trainable parameters, this framework demonstrates outstanding multi-modal processing capabilities, particularly excelling in tasks such as text grounding and text spotting, consistently surpassing traditional MLLMs.
\end{abstract}

\section{Introduction}

Optical Character Recognition (OCR) is a critical technology in the field of computer vision, encompassing tasks such as text spotting, text grounding, and Visual Question Answering (VQA). OCR plays a significant role in scenarios such as document digitization, intelligent transportation, and multi-modal interaction. Position prediction allows the alignment of recognition results with their corresponding locations in the image, thereby enabling interactive document editing, and assisting in the review of machine recognition results. Text spotting specialist models can output both the position and recognition results, thus addressing challenges such as complex backgrounds and various fonts, which has led to the development of various specialist models (e.g., SPTS~\cite{peng2022spts}, InstructOCR~\cite{duan2024instructocr}).

Driven by breakthroughs in Multi-modal Large Language Models (MLLMs) such as ~\cite{li2023blip,liu2024visual,bai2023qwen,Monkey}, recent research has accelerated the adoption of MLLM-based frameworks in OCR to address long-standing challenges like VQA~\cite{yu2024texthawk,feng2023docpedia}, text spotting~\cite{liu2024textmonkey} and text grounding~\cite{mplug-docowl1.5}.
These MLLMs leverage pre-trained visual backbones to extract powerful visual features, which are then fed into the LLM through modality connectors. The LLM utilizes its robust understanding capabilities to perform semantic reasoning and generate the final answer. By training on large-scale image-text data, MLLMs can capture the complex relationships between images and text, enabling them to excel in a wide range of visual tasks.

Despite the remarkable performance of MLLMs in diverse tasks, they encounter two notable limitations:(1) Their reliance on Large Language Models (LLMs) as decoders, which are inherently tailored for linguistic processing, limits their ability to perform the positional reasoning essential for precise visual tasks such as coordinate predictions and text localization.
(2) MLLMs require extensive computational resources and large-scale image-text data due to their substantial trainable parameter sizes. In contrast, specialist models achieve impressive performance with fewer parameters and effectively handle tasks like coordinate prediction and text recognition, areas where MLLMs often lag behind.

\begin{figure}
  \centering
  \includegraphics[width=0.99\linewidth]{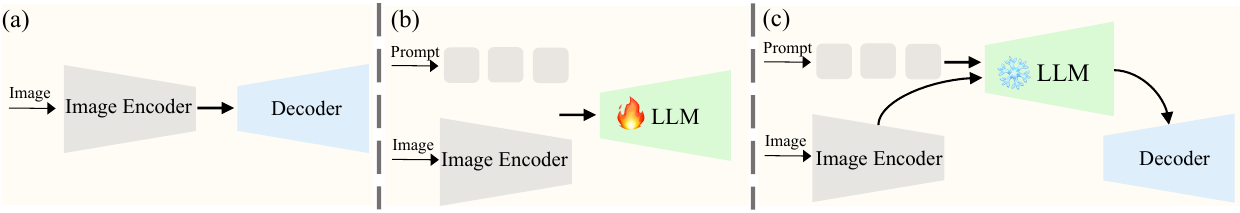}
  \caption{(a) Represents the specialist model for text spotting, which consists of an image encoder and decoder. The model can only output the text within the entire image along with its corresponding positions. (b) Represents mainstream MLLMs, where the image encoder extracts visual features and the LLM completes various multi-modal tasks. (c) Our proposed method, PositionOCR, employs a text spotting model that is guided by the LLM to accomplish various multi-modal tasks.}
  \label{fig:int}
\end{figure}

In this paper, we propose PositionOCR, which demonstrates exceptional capabilities in position reasoning while also performing other multi-modal tasks, such as VQA, without requiring the training of the LLM. We design the LLM to serve as the central processing unit, responsible for processing the input prompts and image features, as shown in Figure \ref{fig:int} (c). This approach allows for effective integration of language understanding with visual information, enabling the model to demonstrate superior performance in various tasks. Notably, it demonstrates exceptional performance in position-related tasks, such as text grounding.

The training of PositionOCR consists of two stages: obtaining a specialist model and following instructions. In the first stage, we develop a specialist model focused on text spotting. This is an image-to-sequence approach that requires the model to recognize the text in images and perceive the specific position of the text within the image, thereby achieving the alignment of text and position. Through this approach, the specialist model learns the transformation relationship from image to sequence, extracting serialized text data from visual information, thereby improving its understanding of the alignment between images and text.

Upon acquiring the specialist model, we first align the specialist model with the LLM using the tasks from the first phase, inputting prompts and images into the LLM, and establishing a connection between the LLM and the decoder through a connector. Subsequently, we train the model using substantial instructional data, enabling the specialist model to understand and execute various downstream tasks. This data includes instructions for different tasks as well as corresponding input-output examples.

In summary, the main contributions are three-fold:
\begin{enumerate}
\item We introduce a novel multi-modal framework, PositionOCR, which integrates the strengths of specialist models with the contextual capabilities of Large Language Models (LLMs). This hybrid approach excels in tasks commonly associated with multi-modal processing, such as text spotting and text grounding
\item Our methodology eradicates the need for extensive LLM training by fine-tuning the specialist models with targeted instructional data. This results in a remarkably efficient architecture, comprising only 131 million trainable parameters, capable of executing multi-modal tasks across various domains with high efficacy.
\item We present comprehensive experimental evidence from public datasets, demonstrating that PositionOCR achieves outstanding performance and surpasses existing solutions across a variety of downstream tasks. It has achieved state-of-the-art results specifically in text grounding and text spotting, further highlighting its effectiveness and competitiveness compared with results delivered by traditional MLLMs.
\end{enumerate}

\section{Related work}

\subsection{Multi-modal large language models}
Large Language Models (LLMs)~\cite{achiam2023gpt,bai2023qwen,touvron2023llama,brown2020language,zhang2022opt} have achieved significant success in tasks such as language understanding and generation, prompting researchers to integrate visual encoders into LLMs for visual language understanding. These Multi-modal Large Language Models (MLLMs) approaches use modality connectors to align the semantic spaces of visual features with the language features in LLMs. For example, Flamingo~\cite{alayrac2022flamingo} integrates a Perceiver Resampler to enhance visual representation, while BLIP2~\cite{li2023blip} employs a Q-Former to connect visual features with the LLM.
General-purpose MLLMs emphasize task generalization, whereas OCR tasks place greater importance on resolution and corresponding training data. 

Monkey~\cite{Monkey} divides input images into uniform patches and supports resolutions up to 1344×896 pixels, which allows for a more detailed capture of visuals. It also employs a multi-level description generation method to enrich the context.
mPLUG-Owl\cite{ye2023mplug} introduces a training paradigm for large language models through modularization and constructs an instruction evaluation set to assess the capabilities of different models in the context of visual-related tasks. Building on this foundation, 
Vary\cite{wei2025vary} achieves dense and fine-grained vision perception by scaling up the vision vocabulary of LVLMs.
Some MLLMs focus on OCR tasks, and this specialization allows them to perform exceptionally well in the field of text recognition. Textmonkey~\cite{liu2024textmonkey} adopts Shifted Window Attention to expand the input resolutions and can output positions. UReader~\cite{ye2023ureader} designs a shape-adaptive cropping module to process high-resolution images. DocPedia~\cite{feng2023docpedia} processes visual input in the frequency domain rather than the pixel space to capture a greater amount of visual and textual information. mPLUG-DocOwl~\cite{ye2023mplug} and mPLUG-DocOwl1.5~\cite{mplug-docowl1.5} are MLLMs that focus on OCR tasks and further enhance performance through Unified Structure Learning.

\subsection{Sequence-based specialist model}
Text spotting aims to detect and recognize characters directly within images. Sequence-based methods use transformers to predict sequences to handle various tasks. Pix2seq V1 and V2~\cite{chen2021pix2seq, chen2022unified} output bounding boxes and labels as discrete token sequences, achieving competitive results on the challenging COCO dataset. Inspired by Pix2seq, the SPTS series~\cite{peng2022spts,liu2023spts} uses the central point of text regions to represent the position and auto-regressively predicts coordinate tokens and word transcription tokens. Donut~\cite{kim2022ocr} proposes the first OCR-free document understanding framework based on Transformers. UNITS~\cite{kil2023towards} unifies various detection formats (including points, quadrilaterals, and polygons) into a sequence generation paradigm. OmniParser~\cite{wan2024omniparser} is capable of handling text recognition, key information extraction, and table recognition simultaneously using a single model. InstructOCR~\cite{duan2024instructocr} introduces instructions to enhance the recognition performance of text spotting.

Text spotting requires the model to recognize text in images and perceive the specific position of the text within the image. In this way, the model learns the transformation relationship from images to sequences, allowing for a better understanding of the alignment between images and text. The LLM possesses strong comprehension capabilities, and by combining text spotting with LLM, it is possible to obtain a model that aligns images and text. This model can interact with humans through natural language via instruction tuning.

\begin{figure*}[t]
  \centering
  \includegraphics[width=0.98\textwidth]{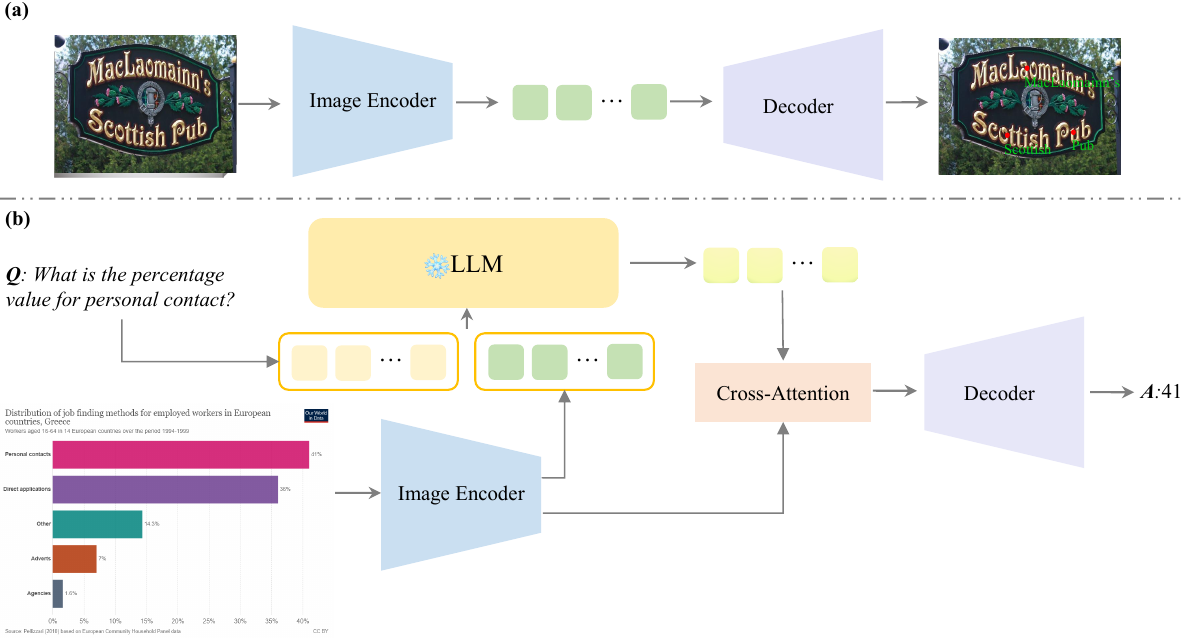}
  \caption{Overall framework of PositionOCR. In the first stage, a specialist model is developed, proficient in performing detection and recognition tasks. In the second stage, a Large Language Model (LLM) is introduced to achieve alignment between the two components using data from text spotting, followed by instruction tuning to enable interaction through human language.}
  \label{fig:framework}
\end{figure*}

\section{Method}
We introduce PositionOCR, an end-to-end document understanding framework. The overall structure is shown in Figure~\ref{fig:framework}.
In the following sections, we will detail the structure of PositionOCR.

\subsection{Architecture}

\textbf{Image encoder}. The image encoder employs the ResNet50~\cite{he2016deep} architecture to extract features from the input image, initialized with weights from the ODM~\cite{duan2024odm} model.

\textbf{Large language model}.
The LLM employed is the Qwen2.5~\cite{qwen2025qwen25technicalreport} series, specifically the 7B variant. This model has undergone pretraining with over 18 trillion tokens and has been fine-tuned through supervised learning with more than 1 million examples, along with multi-stage reinforcement learning optimization. The input text, combined with visual features, is processed as a unified input through Qwen2.5-7B, resulting in embedding vectors. These vectors are then utilized for subsequent reasoning tasks, enhancing the model's capability to integrate and process both textual and visual information.

\textbf{Decoder}. The decoder in PositionOCR employs an auto-regressive Transformer to generate extended sequences for all text instances. In the context of text spotting, we introduce a separator token $<sep>$ to differentiate between instances. Each text instance is represented by a sequence comprising four components: $[x,y,t,<sep>]$. Here, $(x,y)$ denotes the center point coordinates, which are uniformly discretized into integers ranging from 1 to 1000. The component $t$ represents the transcribed text, while $<sep>$ acts as a special separator token to mark the end of an instance. For the VQA task, the answer is simply represented as a direct sequence.

The decoding process leverages the hidden state $\mathbf{h}_t$ of the Transformer decoder and the vocabulary projection matrix $\mathbf{W}_d$ to estimate the probability of the next token in the sequence. This is articulated mathematically as follows:
\begin{equation}
P(\mathbf{y}_t | \mathbf{y}_{1:t-1}, \mathbf{H}_v) = \text{Softmax}(\mathbf{W}_d \mathbf{h}_t)
\end{equation}
Here, $\mathbf{y}_t$ represents the token predicted at the current time step, $\mathbf{y}_{1:t-1}$ denotes the sequence of previously generated tokens, and $\mathbf{H}_v$ encapsulates the encoded visual information.

\subsection{Training strategy}
The training strategy of PositionOCR encompasses two key components: acquiring a specialist model and instruction tuning to enable interaction with human language. 

\textbf{Specialist model}. The specialist model, which follows an image-to-sequence architecture, has achieved competitive results in the text spotting task. As shown in Figure ~\ref{fig:framework}, this architecture consists of an image encoder and a decoder. The image encoder extracts visual features, while the decoder outputs the positions and corresponding recognition results.
To demonstrate the multi-task capabilities of PositionOCR, document data and object detection data were incorporated during the training process of the specialist model. The training process consists of two steps: first, training the model's image-to-sequence capability using data from text spotting. Subsequently, the LLM is integrated to continue training with the text spotting data. For the text spotting task, the input prompt is: ``\textit{Recognize all the text in the image and provide the exact coordinates of each detected text block}''. In contrast, for the object detection task, the input prompt is: ``\textit{Please perform object detection in the image and provide the coordinates of the bounding boxes for each detected object}''.

Recognize all the text in the image and provide the exact coordinates of each detected text block
\textbf{Instruction tuning}. In this stage, the specialist model is endowed with multi-modal understanding capabilities, with instructions input into the LLM and answers output by the specialist model. To demonstrate the performance of PositionOCR across multiple tasks, we collect publicly available datasets from various scenes to train the model's instruction-tuning capability using a substantial amount of instructional data. Additionally, our framework is composed of a specialist model, which inherently enables the execution of position-related tasks, such as text grounding, allowing the model to output positions through instructions.

\subsection{Loss function}
In PositionOCR, the training objective is to predict tokens, and we utilize the standard cross-entropy loss for model training. This loss function aims to maximize the likelihood of the correct tokens during training. The mathematical expression for the cross-entropy loss is as follows:
\begin{equation}
\mathcal{L}_{seq} = \text{maximize} \sum_{i=1}^{L} w_i \log P(\tilde{s}_i | I, s_{1:i})
\end{equation}
where $I$ is the input image, $s$ is the input sequence, $\tilde{s}$ is the output sequence, $L$ is the length of the sequence, and $w_i$ is the weight of the likelihood of the $i-th$ token, which is empirically set to 1.

\section{Experiment}

\subsection{Datasets}

\begin{table}[h]
    \centering
    \caption{Details of datasets used in specialist model training.} 
    \begin{tabular}{c|c|l}
    \toprule
    Task & Samples & Datasets \\
    \midrule
        \multirow{1}{*}{\begin{tabular}[c]{@{}c@{}}Document\\ \end{tabular}} & \multirow{1}{*}{1344.5k} & IIT-CDIP~\cite{iit}, DocVQA\cite{mathew2021docvqa}, InfoVQA\cite{mathew2022infographicvqa} \\
    \midrule
        \multirow{1}{*}{\begin{tabular}[c]{@{}c@{}}Chart\\ \end{tabular}} & \multirow{1}{*}{18.3k} & ChartQA\cite{masry2022chartqa} \\
     \midrule
        \multirow{3}{*}{\begin{tabular}[c]{@{}c@{}}Scene text\\ \end{tabular}} & \multirow{3}{*}{387.1k} & Total-Text\cite{ch2017total}, TextOCR~\cite{singh2021textocr}, ICDAR13~\cite{karatzas2013icdar}, ICDAR15~\cite{karatzas2015icdar}\\
        & & ICDAR2017 MLT\cite{nayef2017icdar2017}, HierText~\cite{long2022towards}, OpenVINO~\cite{krylov2021open} \\
        & & SCUT-CTW1500\cite{yuliang2017detecting} , Curve Synthetic Dataset\cite{liu2020abcnet}  \\
    \midrule
    \multirow{1}{*}{Obj. detection} & \multirow{1}{*}{334.3k} & COCO~\cite{lin2015microsoftcococommonobjects}, Objects365~\cite{shao2019objects365} \\
    \midrule
    Total & \multicolumn{2}{c}{2084.2k} \\
    \bottomrule
    \end{tabular}
    \label{tab:stage1}
\end{table}

\begin{table*}[h]
    \centering
    \vspace{\baselineskip} 
    \caption{Details of datasets used in instruction tuning.} 
    \vspace{\baselineskip} 
    \begin{tabular}{c|c|l}
    \toprule
    Domain & Samples & Datasets \\
    \midrule
        \multirow{3}{*}{\begin{tabular}[c]{@{}c@{}}Document\end{tabular}} & \multirow{3}{*}{4758.8k} &  DocGenome~\cite{xia2024docgenome}, DocVQA~\cite{mathew2021docvqa}, InfoVQA~\cite{mathew2022infographicvqa}, KLC~\cite{stanislawek2021kleister} \\
        & & VisualMRC~\cite{tanaka2021visualmrc}, DocReason25k~\cite{mplug-docowl1.5}, Sujet-Finance~\cite{sujet-finance-instruct-177k},  \\
        & & POIE~\cite{kuang2023visual}, SROIE~\cite{huang2019icdar2019sroie}, FUNSD~\cite{jaume2019funsd}, TextCaps~\cite{sidorov2020textcaps}, \\
        & &  SynthDoG-en~\cite{kim2022ocr}, Docmatix~\cite{laurenccon2024building}, OCR-VQA~\cite{8978122}\\
    \midrule
        \multirow{2}{*}{\begin{tabular}[c]{@{}c@{}}Table\end{tabular}} & \multirow{2}{*}{270.9k} &WikiTableQuestions~\cite{pasupat2015compositional},  TabMWP~\cite{lu2023dynamic}, DeepForm~\cite{svetlichnaya2020deepform}, \\
        & & TableFact~\cite{chen2019tabfact},TableBench~\cite{wu2024tablebench}, TableVQA~\cite{kim2024tablevqa} \\
    \midrule
        \multirow{1}{*}{\begin{tabular}[c]{@{}c@{}}Chart\end{tabular}} & \multirow{1}{*}{4083.5k} & ChartQA~\cite{masry2022chartqa}, FigureQA~\cite{figqa}, UniChart~\cite{masry2023unichart}, ChartBench~\cite{xu2023chartbench}, \\
        & & DVQA~\cite{kafle2018dvqa}, ai2d~\cite{hiippala2021ai2d} \\        
    \midrule
        \multirow{2}{*}{\begin{tabular}[c]{@{}c@{}}Formula\end{tabular}} & \multirow{2}{*}{1675.1k} & UniMER-1M~\cite{wang2024unimernet}, CROHME 2014~\cite{mouchere2014icfhr}, CROHME 2016~\cite{mouchere2016icfhr2016},  \\
        & & CROHME 2019~\cite{mahdavi2019icdar}, Latex-OCR~\cite{latex-ocr}, IAM Handwriting~\cite{marti2002iam}, \\
        & & HME100k ~\cite{yuan2022syntax}
        \\
    \midrule
        \multirow{3}{*}{\begin{tabular}[c]{@{}c@{}}Scene text\end{tabular}} & \multirow{3}{*}{469.6k} & ICDAR2017 MLT~\cite{nayef2017icdar2017}, Curved Synthetic Dataset 150k~\cite{liu2020abcnet},    \\
        & &  TextOCR~\cite{singh2021textocr}, HierText~\cite{long2022towards}, ST-VQA~\cite{biten2019scene}, TextOCR~\cite{singh2021textocr},  \\        
        & & Total-Text~\cite{ch2017total}, ICDAR2013~\cite{karatzas2013icdar}, ICDAR2015~\cite{karatzas2015icdar},   \\
        & &  TextVQA~\cite{singh2019towardstextvqa}, OpenVINO~\cite{krylov2021open},
         \\
    \midrule
    \multirow{1}{*}{Obj. detection} & \multirow{1}{*}{117.2k} & COCO~\cite{lin2015microsoftcococommonobjects} \\
    \midrule
        \multirow{1}{*}{\begin{tabular}[c]{@{}c@{}}Text grounding\end{tabular}} & \multirow{1}{*}{800k} & DocStruct4M-subset~\cite{mplug-docowl1.5}
   \\    
    \midrule
        \multirow{1}{*}{\begin{tabular}[c]{@{}c@{}}Obj. grounding\end{tabular}} & \multirow{1}{*}{429.9k} & RefClef~\cite{luo2017comprehension}, RefCOCO~\cite{luo2017comprehension}, RefCOCO+~\cite{luo2017comprehension}, RefCOCOg~\cite{luo2017comprehension} 
   \\ 
    \midrule
    Total & \multicolumn{2}{c}{12.6M} \\
    \bottomrule
    \end{tabular}
    \label{tab:stage2}
\end{table*}

\textbf{Specialist model}. In this training stage, we use text spotting data from both documents and natural scenes, with a total training dataset of 2.1M. The datasets are shown in Table~\ref{tab:stage1}. Specifically, for the document data, we randomly sample 1.3M images from the IIT-CDIP~\cite{iit} dataset and employ PPOCRv3\cite{li2022pp} to generate pseudo labels (i.e., text and position in the image). To further demonstrate the capabilities of PositionOCR, we incorporate object detection data from COCO and Objects365. Given that the Objects365 dataset contains 2 million images, this is substantial for us, so we randomly select 217k images from Objects365.


\textbf{Instruction tuning}. In this training stage, we utilize a diverse set of datasets to enhance the model's ability to understand and execute instructions across various domains, with a total training dataset of 12.6M. The datasets are shown in Table~\ref{tab:stage2}. 
After being trained on multi-domain data, the model gains the ability to accept instructions in human language. We then fine-tune the model further using the training sets of downstream tasks. Additionally, we utilize another dataset of 3.1M samples for training during this stage.
These include document datasets such as DocVQA, InfoVQA, DeepForm, OCR-VQA, KLC, and VisualMRC. Table datasets such as TableFact and WikiTableQuestions. Chart datasets include ChartQA. Natural scene datasets include TextVQA, ST-VQA, ICDAR2013, ICDAR2015, Total-Text, TextOCR, Curved Synthetic Dataset 150k, MLT-2017, HierText, TextCaps. Text grounding datasets include DocStruct4M-subset. Object grounding datasets include RefClef, RefCOCO, RefCOCO+ and RefCOCOg. KIE datasets include FUNSD, POIE and SROIE.

\subsection{Implementation details}
The entire model is distributively trained on 32 NVIDIA A100-80G GPUs. During the training process in the specialist model, to enhance training efficiency, the short side of the input image is randomly resized to a range from 704 to 1024 (intervals of 32), and the maximum length of the image is set to 1024. The batch size per GPU is 18, and the model is trained for 130 epochs, with an initial 5-epoch warm-up phase. We use the AdamW optimizer with a learning rate of $4.6 \times 10^{-4}$. Subsequently, the model is trained for another $40$ epochs, with a fixed learning rate of $6 \times 10^{-5}$, and the maximum length of the image is to $1824$. Then, the LLM is integrated into the specialist model, and the model is further trained for another $40$ epochs.
For instruction tuning, we first fine-tune the model for $10$ epochs using all the instructions data. And the maximum length of the image is set to 1024. Subsequently, the model is trained for another $5$ epochs, and the maximum length of image is set as $1728$. The model is then fine-tuned using downstream data, with training conducted for 10 epochs during this phase.

\subsection{Comparison with text grounding results}
Text grounding aims to associate language with its related real-world information, necessitating an understanding of human language alongside the capability to output the corresponding positions. We test the model's performance using the DocLocal4K dataset proposed by  DocOwl-1.5~\cite{mplug-docowl1.5} without fine-tuning. This dataset contains 4,250 samples, with text granularity levels including ``Word'', ``Phrase'', ``Line'', and ``Block''. The IOU@0.5 is used to evaluate the text grounding performance.

\begin{figure*}[t]
  \centering
  \includegraphics[width=0.98\textwidth]{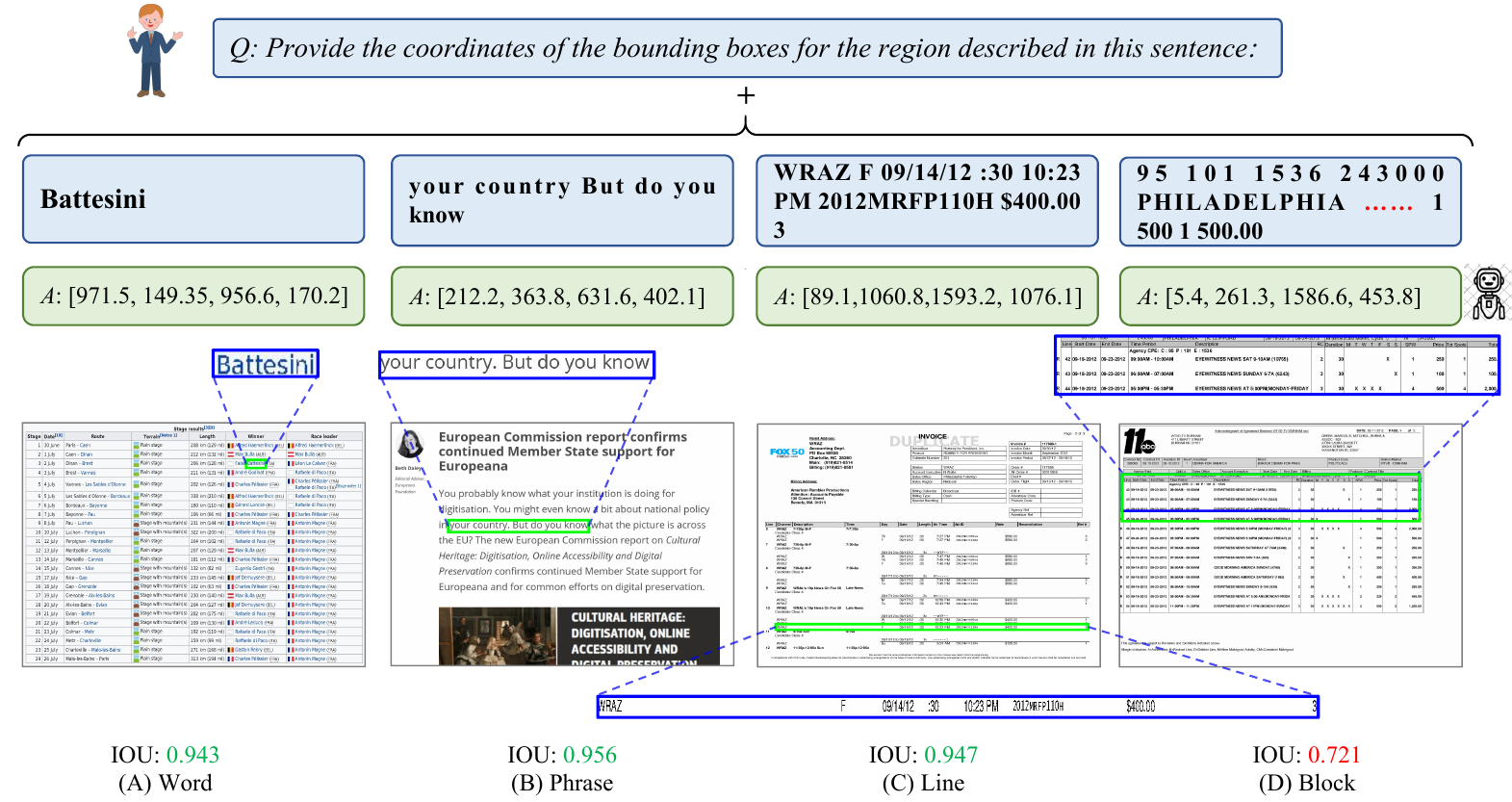}
  \caption{PositionOCR's visualization results on text granularity include `Word'', ``Phrase'', ``Line'', and ``Block''.}
  \label{fig:textgrounding}
\end{figure*}

Table~\ref{tab:result_testgrounding} shows the results for text grounding. Compared to  DOGE, our method achieves a metric of 83.0\%, which notably exceeds the performance benchmark of 0.4\%. This improvement demonstrates the effectiveness of PositionOCR in the context of text grounding. Notably, the text granularity is particularly evident at the ``Word'' level, where  DOGE achieves only 74.7\%, whereas PositionOCR attains 84.0\%, surpassing it by 9.3\%. This indicates that our model performs better at fine granularity, which relates to the specialist model making ``Word'' level predictions. As the granularity of the target boxes increases, our advantage decreases; at the ``Block'' level, our performance is weaker than that of DOGE. This also suggests that MLLMs are more adept at making coarse-grained predictions, while their perception of fine granularity is relatively weak~\cite{zhang2025mllms}. The visualizations at different levels are shown in Figure~\ref{fig:textgrounding}.

\begin{table}[h]
    \centering
  \caption{The performance of different methods on the text localization evaluation set DocLocal4K. ``TP'' refers to Trainable Parameters. The IOU@0.5 is used to evaluate the text grounding performance}
  \begin{tabular}{>{\centering\arraybackslash}p{0.2\linewidth}%
                  >{\centering\arraybackslash}p{0.08\linewidth}%
                  >{\centering\arraybackslash}p{0.08\linewidth}%
                  >{\centering\arraybackslash}p{0.08\linewidth}%
                  >{\centering\arraybackslash}p{0.08\linewidth}%
                  >{\centering\arraybackslash}p{0.08\linewidth}%
                  >{\centering\arraybackslash}p{0.08\linewidth}}
    \toprule
    \multirow{2}{*}{Methods}& \multirow{2}{*}{TP}  & \multicolumn{5}{c}{Text Grounding}  \\
    \cmidrule(lr){3-7} 
     & & Block & Line  & Phrase & Word & All \\
    \midrule
    Qwen2.5-VL~\cite{bai2025qwen2} &7B& 0.3 &32.2   & 6.2 & 20.6 & 15.0 \\
    DocOwl-1.5~\cite{liu2024textmonkey} &8.1B&91.3 & 85.9   & 76.4& 70.4& 80.4 \\
    DOGE~\cite{zhou2024doge} &8B&95.4 & 84.1   & 78.4& 74.7& 82.6 \\
    \midrule
    PositionOCR &131M& 76.5  & \textbf{89.1}  & \textbf{81.9}  & \textbf{84.0}   &\textbf{83.0}  \\
    \bottomrule
  \end{tabular}
    \label{tab:result_testgrounding}
\end{table}

To further evaluate the model's performance on grounding tasks, we construct a new dataset derived from TextVQA~\cite{singh2019towardstextvqa}, termed TextVQA-G (TextVQA-Grounding). We utilized the validation set of TextVQA as our test set. In this dataset, while the original questions are retained, the target responses are converted into localization coordinates. The dataset provides two types of annotation formats: quad (four-point representation) and box (two-point representation). In this work, we use the box format. For instance, given the query ``which edition is this book?'', the ground truth is represented as [221, 256, 304, 283]. The resulting dataset comprises 25,808 questions in the training set and 3,536 questions in the test set. 

We conducted fine-tuning experiments on both PositionOCR and Qwen2.5-VL using this dataset. The comparative results are presented in Table~\ref{tab:textvqa_grounding}. As shown in the table, PositionOCR achieves an accuracy of 80.31\%, slightly outperforming Qwen2.5-VL, which scores 79.92\%.

\begin{table}[h]
    \centering
    \caption{Performance comparison on the TextVQA-G dataset. The metric used is Accuracy @ IoU=0.5.}
    \begin{tabular}{lcc}
    \toprule
    Methods & Trainable Params & Accuracy @ IoU=0.5 \\
    \midrule
    Qwen2.5-VL~\cite{bai2025qwen2} & 7B & 79.92\% \\
    PositionOCR & 131M & \textbf{80.31\%} \\
    \bottomrule
    \end{tabular}
    \label{tab:textvqa_grounding}
\end{table}

\subsection{Comparison with text spotting results}
To further demonstrate the position prediction capabilities of PositionOCR, we evaluate its performance on the text spotting datasets without requiring fine-tuning. During the inference phase, the prompt input for the text encoder is ``\textit{Recognize all the text in the image and provide the exact coordinates of each detected text block}''. The maximum length of the images is less than 1728 pixels, and the minimum length is 1024 pixels. We use the point-based metrics proposed in SPTS to evaluate the model. Specifically, ICDAR2015 is a multi-directional text dataset, while Total-Text is a dataset consisting of text in arbitrary shapes.

Table~\ref{tab:text_spotting_results} shows the results for text spotting. Compared to TextMonkey, we surpass it by 5.2\% on Total-Text and by 22.6\% on ICDAR2015. In comparison to the 9.7B model, our lightweight model achieves superior performance. This demonstrates the advantages of our approach in position awareness.

\begin{table}[h]
  \caption{Text spotting results on Total-Text and ICDAR2015 in the VQA task. ‘None’ means lexicon-free. ‘Full’ indicates that we use all the words that appeared in the test set. ‘S’, ‘W’, and ‘G’ represent recognition with ‘Strong’, ‘Weak’, and ‘Generic’ lexicons, respectively. ``TP'' refers to the number of trainable parameters in the model.}
  \centering
  \begin{tabular}{>{\centering\arraybackslash}p{0.2\linewidth}%
                  >{\centering\arraybackslash}p{0.08\linewidth}%
                  >{\centering\arraybackslash}p{0.08\linewidth}%
                  >{\centering\arraybackslash}p{0.08\linewidth}%
                  >{\centering\arraybackslash}p{0.08\linewidth}%
                  >{\centering\arraybackslash}p{0.08\linewidth}%
                  >{\centering\arraybackslash}p{0.08\linewidth}}
    \toprule
    \multirow{2}{*}{Methods}& \multirow{2}{*}{TP} & \multicolumn{2}{c}{Total-Text}  & \multicolumn{3}{c}{ICDAR2015}  \\
    \cmidrule(lr){3-4} \cmidrule(lr){5-7}
     & & None & Full  & S & W & G \\
    \midrule
    Qwen2.5-VL~\cite{bai2025qwen2} &7B& 45.2 & 51.1   & 46.4 & 43.7 & 42.3 \\
    TextMonkey~\cite{liu2024textmonkey} &9.7B&61.4 & -   & -& -& 45.1 \\
    \midrule
    PositionOCR &131M& \textbf{66.6} & \textbf{77.4}  & \textbf{74.1} & \textbf{71.1}  &\textbf{67.7} \\
    \bottomrule
  \end{tabular}
\label{tab:text_spotting_results}
\end{table}

\subsection{Comparison with results on document benchmarks}
PositionOCR is capable of performing VQA tasks across various contexts, including documents, charts, and tables. We compare our approach with recent MLLMs methods for document understanding. During the inference phase, the maximum input size is set to 1728 pixels, while the minimum input size is set to 1280 pixels. As shown in Table \ref{tab:result1}, our method achieves 69.8\% on the DocVQA dataset, surpassing MLLMs methods such as DocPeida, DocOwl, UReader, Monkey, and Qwen-vl. Our approach also delivers competitive results on other datasets, with performance in the DeepForm dataset ranking only behind DocOwl-1.5. Our method performs just slightly below Mini-Monkey on the SROIE dataset while achieving state-of-the-art performance on the POIE dataset. We also include traditional document understanding methods, such as Dessurt and Donut. Notably, Donut's pre-trained images even reach 11M, yet our method surpasses all results from these methods.

\begin{table*}[h]
    \centering
    \caption{Comparison with MLLMs and OCR-free methods on various types of image understanding tasks. All evaluation benchmarks use the officially designated metrics. ``TP'' refers to the number of trainable parameters in the model.}
    \vspace{\baselineskip} 
    \begin{tabular}{c|c|cccccc}
        \toprule 
        Methods & TP  & DocVQA &  DeepForm &  ChartQA  & TabFact &SROIE &POIE\\
        \midrule
        Dessurt~\cite{davis2022end} & 127M & 63.2 &   - & -   & -  & -  & -\\         
        Donut~\cite{kim2022ocr} & 176M & 67.5  & 61.6  & 41.8  & 54.6 & -   & -\\
        \midrule
        DocPeida~\cite{feng2023docpedia} & 7.1B  & 47.1   & - & 46.9   & -  & 21.4  & 39.9\\
        DocOwl~\cite{ye2023mplug} & 7.3B  & 62.2  & 42.6 &  57.4  & 67.6 & 1.7 & 2.5 \\
        UReader~\cite{UReader} & 7.1B  & 65.4  & 49.5 & 59.3  & 67.6 & -  & -\\  
        QwenVL~\cite{bai2023qwen} & 9.6B  & 65.1   & 65.7  & -  & - & - \\ 
        Monkey~\cite{Monkey} & 9.8B  & 66.5 & 40.6 & 65.1 & -  & 41.9 & 19.9\\
        DocKylin~\cite{zhang2024dockylin} & 7.1B & 77.3 & -  & 66.8  & - & - & - \\ 
        Mini-Monkey~\cite{MiNi-Monkey} & 2B & 87.4 & -  & 76.5  & - & 70.3  & 69.9\\
        DocOwl-1.5~\cite{mplug-docowl1.5} & 8.1B & 81.6 & 68.8  & 70.5  & 80.4 & - & -\\
        TextMonkey~\cite{liu2024textmonkey} & 9.7B & 73.0  & -  & 66.9  & - & 47.0  & 27.9\\
        HRVDA~\cite{liu2024hrvda} & 7.1B  & 72.1 & 63.2  & 67.6  & 72.3 & - & - \\
        \midrule
        PositionOCR & 131M   & 69.8 & 66.3  &58.8  & 62.8 & 63.8 & \textbf{77.5}\\     
        \bottomrule
    \end{tabular}
    \label{tab:result1}
\end{table*}

\subsection{Cross-Task generalization analysis}
The specialist model of PositionOCR is based on an image-to-sequence architecture, which means that this specialist model can also perform object detection tasks. Therefore, we incorporated object detection data into the training of the specialist model, specifically from Objects365 and COCO. 
In terms of object grounding metrics, as shown in Table \ref{tab:gr_results}, despite using a very small amount of data—only 334k images for pretraining, our approach remains competitive in object grounding metrics. Compared to traditional specialist models, such as Speaker, our method even demonstrates superior performance. However, when compared to the latest MLLMs-based methods, such as Shikra, our metrics are lower, as they leverage a significantly larger amount of data, which we are unable to match.

\begin{table*}[h]
  \centering
  \caption{Results of object grounding on Refclef, Refcoco, Refcoco+, and Refcocog.}
  \vspace{\baselineskip} 
  \begin{tabular}{c|cc|ccc|ccc|cc}
    \toprule
    \multirow{2}{*}{Methods}  & \multicolumn{2}{c}{Refclef}  & \multicolumn{3}{c}{Refcoco} & \multicolumn{3}{c}{Refcoco+} & \multicolumn{2}{c}{Refcocog} \\
    \cmidrule(lr){2-3} \cmidrule(lr){4-6}  \cmidrule(lr){7-9} \cmidrule(lr){10-11}
     &  Test  & Val & TestA & TestB  & Val & TestA & TestB  & Val & Test & Val \\
    \midrule
    CITE~\cite{plummer2018conditional}  &34.13  & - & - & - & - & - &-&- & - & - \\
    MMI~\cite{mao2016generation}  & - &- &64.9 &54.5 & - & 54.0 & 42.8 & - & - & - \\
    Speaker~\cite{yu2017joint}  & - &- &67.6& 55.1&- &55.8 &43.4& - & - & -\\
    PFOS~\cite{sun2022proposal}&67.9 &-&81.4&73.1&78.4&72.4&55.2&65.8&67.6&67.8\\
    OFA-L*~\cite{wang2022ofa}  &-  & - & 83.6 & 76.3 & 79.9 & 76.0 &61.7&68.2 & 67.5 & 67.5 \\
    Shikra~\cite{chen2023shikra}  &-  &- & 91.1 & 81.8 & 87.8 & 87.7 &74.4&82.8 & 83.1 & 82.6 \\
    \midrule
    PositionOCR & 57.2 &58.9 &73.0& 62.4& 68.3& 58.7 &43.1 &52.4 &58.3& 52.8 \\

    \bottomrule
  \end{tabular}
\label{tab:gr_results}
\end{table*}

\subsection{Ablation Study}
\textbf{Ablation study on text spotting}. 
We propose a framework that endows the specialist model with multi-modal capabilities without the need to train the LLM, while also leveraging the advantages of the specialist model. In this section, we assess the performance of the specialist model. Table~\ref{tab:finetune_ts_results} shows the performance of the specialist model on the text spotting task. Following the training and evaluation protocols for scene text spotting, we fine-tuned the model for 170 epochs on the Total-Text and ICDAR2015 datasets and subsequently evaluated its performance on these datasets.
As observed in Table~\ref{tab:finetune_ts_results}, our model achieves SOTA performance on ICDAR2015 datasets using a generic lexicon, demonstrating the robustness of our pre-training stage, surpassing dedicated models for scene text spotting tasks.

\begin{table}[h]
  \centering
  \caption{The text spotting results of the specialist model on the Total-Text and ICDAR2015 datasets.}
  \begin{tabular}{>{\centering\arraybackslash}p{0.2\linewidth}%
                  >{\centering\arraybackslash}p{0.08\linewidth}%
                  >{\centering\arraybackslash}p{0.08\linewidth}%
                  >{\centering\arraybackslash}p{0.08\linewidth}%
                  >{\centering\arraybackslash}p{0.08\linewidth}%
                  >{\centering\arraybackslash}p{0.08\linewidth}} 
    \toprule
    \multirow{2}{*}{Methods} & \multicolumn{2}{c}{Total-Text}  & \multicolumn{3}{c}{ICDAR2015}  \\
    \cmidrule(lr){2-3} \cmidrule(lr){4-6}
     & None & Full  & S & W & G \\
    \midrule
    TOSS~\cite{tang2022you} & 65.1 & 74.8  & 65.9 & 59.6 & 52.4 \\
    SPTS~\cite{peng2022spts} & 74.2 & 82.4  & 77.5 & 70.2 & 65.8 \\
    SPTS-v2~\cite{liu2023spts} & 75.5 & 84.0 & 82.3 & 77.7 & 72.6 \\
    InstructOCR~\cite{duan2024instructocr} &77.1 &84.1 &82.5 &77.1 &72.1 \\
    \midrule
    PositionOCR & 76.6 & 83.4  & 80.5&\textbf{78.2} &\textbf{74.5} \\
    \bottomrule
  \end{tabular}
\label{tab:finetune_ts_results}
\end{table}

\section{Limitations}
Due to constraints in training resources, PositionOCR encounters limitations in pretraining data and task diversity. Compared to DocOwl-1.5's 4M images and Qwen2.5-VL's 4.1T tokens, PositionOCR merely utilizes 2.1M images. Achieving improved performance requires the utilization of larger and more comprehensive datasets. Additionally, while it includes object detection data, just 334k samples were employed. Future work should incorporate more extensive datasets to enhance generalization and validate applicability across diverse scenarios.

\section{Conclusion}
In this paper, we introduce PositionOCR, a novel hybrid framework that synergizes text spotting specialist model with the contextual reasoning power of LLM, aiming to address the challenges of OCR tasks. Our research reveals that PositionOCR attains exceptional performance, especially in tasks such as text grounding and multi-modal processing, utilizing a mere 131M trainable parameters and surpassing the capabilities of traditional MLLMs. Through rigorous experiments conducted on various public datasets, we confirm the effectiveness of PositionOCR across multiple downstream tasks, emphasizing the indispensable role of specialist models in enhancing multi-modal scenarios. PositionOCR's innovation is rooted in its methodology of fine-tuning the specialist model with instructional data while circumventing the extensive training required by LLMs, thereby achieving notable computational efficiency and improved performance. This study offers fresh perspectives on augmenting the multi-modal capabilities of specialist models and sets the stage for subsequent explorations in this field.





\bibliographystyle{plain} 
\bibliography{ref}

\end{document}